\documentclass[journal]{IEEEtran}

\usepackage{amsmath,amssymb,amsfonts}
\usepackage{mathrsfs}
\usepackage{dsfont}
\usepackage{pifont}         
\usepackage{nicefrac}       

\usepackage[table,xcdraw]{xcolor} 
\usepackage{graphicx}
\usepackage{tikz}

\usepackage{array}
\usepackage{booktabs}       
\usepackage{multirow}       
\usepackage{multicol}       
\usepackage{tabularx}       
\usepackage{makecell}       
\usepackage{arydshln}       

\usepackage{algorithm}
\usepackage{algorithmic}

\usepackage{cite}           
\usepackage{url}
\usepackage{stfloats}       
\usepackage{microtype}      
\usepackage[normalem]{ulem} 

\usepackage{subfig}         
\usepackage{capt-of}        

\usepackage[colorlinks=true, citecolor=green, linkcolor=red, urlcolor=blue]{hyperref}

\definecolor{mygray}{gray}{0.9}
\definecolor{mygray1}{rgb}{0.9, 0.9, 0.9}

\newcommand\hl{\bgroup\markoverwith
	{\textcolor{yellow}{\rule[-.5ex]{2pt}{2.5ex}}}\ULon}

\newcommand{\safeincludegraphics}[2][]{%
  \IfFileExists{#2}{\includegraphics[#1]{#2}}{%
    \fbox{\parbox[c][2.2in][c]{0.95\linewidth}{\centering\scriptsize Missing figure file\\\texttt{\detokenize{#2}}}}%
  }%
}

\allowdisplaybreaks

\begin{document}

\title{ASTRA: Let Arbitrary Subjects Transform in Video Editing}

\author{Fei Shen$^{*}$, Weihao Xu$^{*}$, Rui Yan, Dong Zhang, Xiangbo Shu, \IEEEmembership{Senior Member, IEEE}, \\ Jinhui Tang, \IEEEmembership{Senior Member, IEEE}, and Maocheng Zhao

\thanks{$^{*}$Equal contribution}

\thanks{Fei Shen is with the NExT++ Research Centre, National University of Singapore, Singapore. E-mail: shenfei29@nus.edu.sg.}%
\thanks{Weihao Xu, Rui Yan, Dong Zhang, and Xiangbo Shu are with the School of Computer Science and Engineering, Nanjing University of Science and Technology, Nanjing, China (e-mail: \{whxu, dongzhang, ruiyan, shuxb\}@njust.edu.cn).}

\thanks{Jinhui Tang, and Maocheng Zhao are with the College of Information Science and Technology and
Artificial Intelligence, Nanjing Forestry University, Nanjing 210037, China, (e-mail: \{jinhuitang, mczhao\}@njfu.edu.cn).   (Corresponding author: Jinhui Tang.)}

}

\markboth{IEEE Transactions on Multimedia, Vol. xx, No. xx}
{Shen \MakeLowercase{\textit{et al.}}: ASTRA: Let Arbitrary Subjects Transform in Video Editing}

\maketitle

\begin{abstract}
While recent advances in generative models have propelled video editing, existing methods primarily focus on single or few subjects and struggle in complex, multi-subject scenarios. Specifically, in dense scenes with heavy occlusions, current approaches often suffer from severe mask boundary entanglement and attention dilution, leading to attribute leakage and temporal instability. To address these critical limitations, we present \textbf{an} \textbf{a}rbitrary-\textbf{s}ubjects \textbf{t}raining-free \textbf{r}etargeting and \textbf{a}lignment (ASTRA) framework for video editing that enables an arbitrary number of subjects to transform seamlessly. ASTRA manipulates the appearances of multiple designated subjects while strictly preserving non-target regions, all without requiring any model finetuning or retraining. We achieve this by generating robust multimodal conditioning and precise mask sequences through two core components: a prompt-guided multimodal alignment module and a prior-based mask retargeting module. First, we leverage the comprehensive understanding and generation capabilities of large foundation models to produce multi-subject multimodal information and coherent mask motion sequences. These refined prior mask sequences are then integrated into a pretrained mask-driven video generation model to synthesize the edited video. With its strong generalization capability, ASTRA effectively remedies the issue of insufficient prompt-side multimodal conditioning and overcomes complex mask boundary entanglement in videos featuring varying numbers of subjects, thereby significantly expanding the applicability of video editing. Importantly, ASTRA acts as a versatile plug-in that is completely compatible with diverse mask-driven video generation model, significantly enhancing overall editing performance. Extensive experiments on our newly constructed multi-subject benchmark, MSVBench, verify that ASTRA consistently surpasses state-of-the-art methods. Code, models, and data are available at \textcolor{blue}{\url{https://github.com/XWH-A/ASTRA}}.
\end{abstract}


\maketitle
	

\section{Introduction}
\IEEEPARstart{T}{he} seamless, synchronized transformation of multiple interacting subjects is a captivating visual effect frequently showcased in cinematic and television media~\cite{zuo2026edit,qin2025truncate,donahue2016adversarial}, exemplified by the coordinated team transformations in \textit{Ultraman} and \textit{Sailor Moon}. However, reproducing these compelling multi-subject transformations in real-world videos traditionally relies on specialized equipment and laborious manual character modeling. This reliance fundamentally increases production costs and severely restricts generalization.

\begin{figure}[t]
\centering
\includegraphics[width=0.95\linewidth]{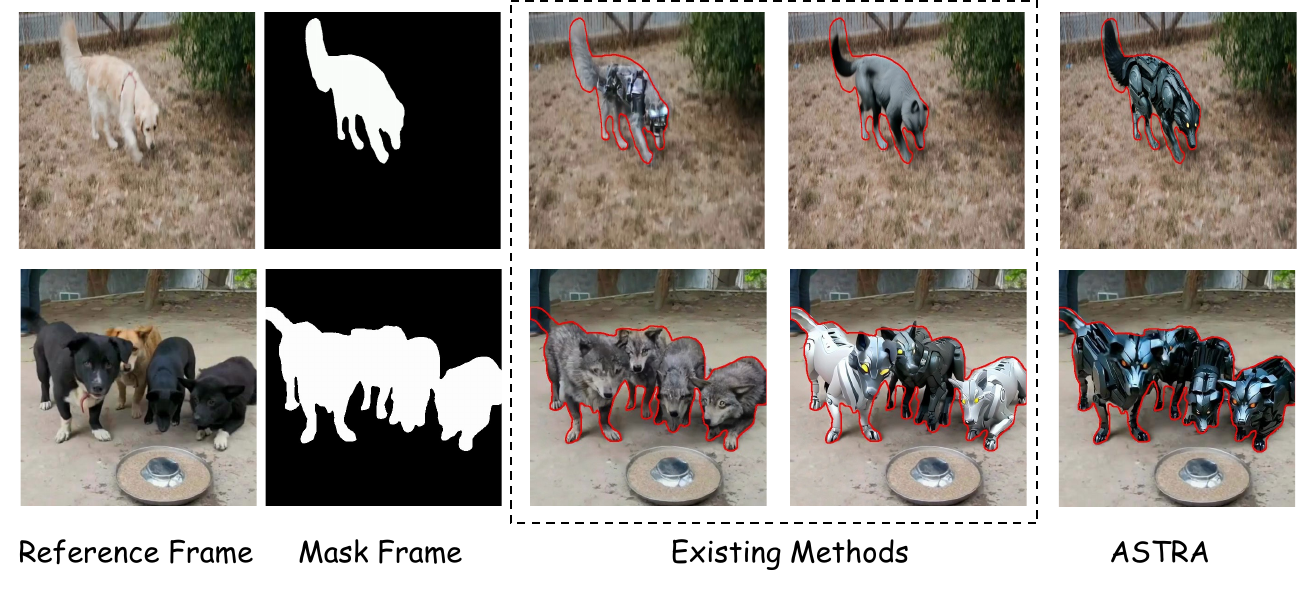}
\vspace{-0.3cm}
\caption{\textbf{Qualitative comparison of complex multi-subject video editing (Dogs $\rightarrow$ Robot Wolves).} Unlike existing methods that struggle with boundary entanglement and attribute leakage, ASTRA delivers precise semantic transformations while strictly preserving non-target regions.}
\label{fig:intro}
\vspace{-0.5cm}
\end{figure}

Driven by rapid advancements in generative models~\cite{gan,ramesh2022hierarchical}, particularly generative adversarial networks~\cite{radford2015unsupervised,odena2017conditional} and diffusion models~\cite{rombach2022high,ramesh2022hierarchical}, video editing~\cite{wu2023tune,ceylan2023pix2video} has witnessed substantial progress. Despite these successes, most existing approaches~\cite{geyer2023tokenflow,wang2025videodirector,ceylan2023pix2video,ku2024anyv2v} remain inherently restricted to single or, at most, paired subjects, often depending heavily on task-specific fine-tuning or rigid guiding masks. This constraint drastically limits their applicability in complex, real-world environments. In dense, multi-subject scenes characterized by intricate layouts and heavy occlusions, current methods frequently exhibit instability and severe perceptual degradation. For instance, as illustrated in the top row of Fig.~\ref{fig:intro}, insufficient multimodal conditioning causes existing models to fail at accurately reflecting the designated editing prompt. 
Furthermore, as depicted in the bottom row of Fig.~\ref{fig:intro}, mask boundary entanglement during segmentation~\cite{ren2016faster,he2017mask} forces edits to bleed across subjects, culminating in catastrophic attribute leakage (e.g., synthesizing a dog's head on a robot wolf's body). 

To mitigate these critical bottlenecks and facilitate universal subject transformations while preserving background integrity, we propose arbitrary-subject training-free retargeting and alignment (ASTRA). As a training-free framework, ASTRA enables the seamless editing of arbitrary subjects in open domain videos without requiring any fine tuning. Unlike prior methodologies that falter under complex spatial layouts, ASTRA delivers robust, high-fidelity edits across subject clusters of various scales. As shown in Fig.~\ref{fig:intro}, by explicitly resolving boundary entanglement and attention dilution, our framework yields temporally consistent and layout-aware transformations even in exceptionally crowded scenes, successfully mitigating the attribute leakage and perceptual decay endemic to existing paradigms.

This performance leap is driven by a synergistic pipeline comprising two core modules integrated into a mask-driven video generation backbone: (i) a prompt-guided multimodal alignment module, and (ii) a prior-based mask retargeting module. In the prompt-guided multimodal alignment phase, we initially isolate the target subjects from the prompt and query a pretrained text-to-image (T2I) model~\cite{rombach2022high,podell2023sdxl,chen2023pixart} to extract a visual prior. By mapping both the editing prompt and this visual prior through a vision-language model (VLM)~\cite{achiam2023gpt,wang2024qwen2,chen2024internvl}, we synthesize robustly aligned multimodal conditions (i.e., augmented textual and visual instructions). Concurrently, our prior-based mask retargeting module tracks per-frame mask state transitions to formulate a temporally coherent mask motion sequence that closely aligns with the source video dynamics. Ultimately, these enriched multimodal conditions and coherent mask sequences are injected into a pretrained mask-driven video generation model to render the final edited synthesis.

Crucially, ASTRA operates as a highly versatile, plug-and-play architecture that is completely compatible with various mask-driven video generator, thereby significantly elevating multi-subject editing baselines. Furthermore, to bridge the critical gap in standardized evaluation for complex video editing, we introduce MSVBench, a comprehensive benchmark comprising 100 challenging sequences that span diverse subject counts, dense inter-subject interactions, and dynamic scene complexities. Extensive qualitative and quantitative evaluations on MSVBench affirm that ASTRA consistently surpasses state-of-the-art baselines. 
Therefore, ASTRA addresses mask-driven text-guided video editing, where designated subjects in an existing source video are transformed while preserving the background and non-target regions.
Our main contributions are summarized as follows:
\begin{itemize}
    \item We propose ASTRA, a training-free video editing framework designed to seamlessly transform an arbitrary number of subjects without additional fine-tuning.
    \item We introduce a prompt-guided multimodal alignment module and a prior-based mask retargeting module. These components generate precise conditions that effectively resolve attention dilution and boundary entanglement in crowded scenes.
    \item We demonstrate that ASTRA functions as a versatile plug-and-play architecture, seamlessly integrating with existing mask-driven video generators to enhance multi-subject editing performance.
    \item We establish MSVBench, a comprehensive benchmark for multi-subject video editing, and show through extensive experiments that ASTRA consistently outperforms state-of-the-art methods.
\end{itemize}


\begin{figure*}[t] 
\begin{center}
\includegraphics[width=\textwidth]{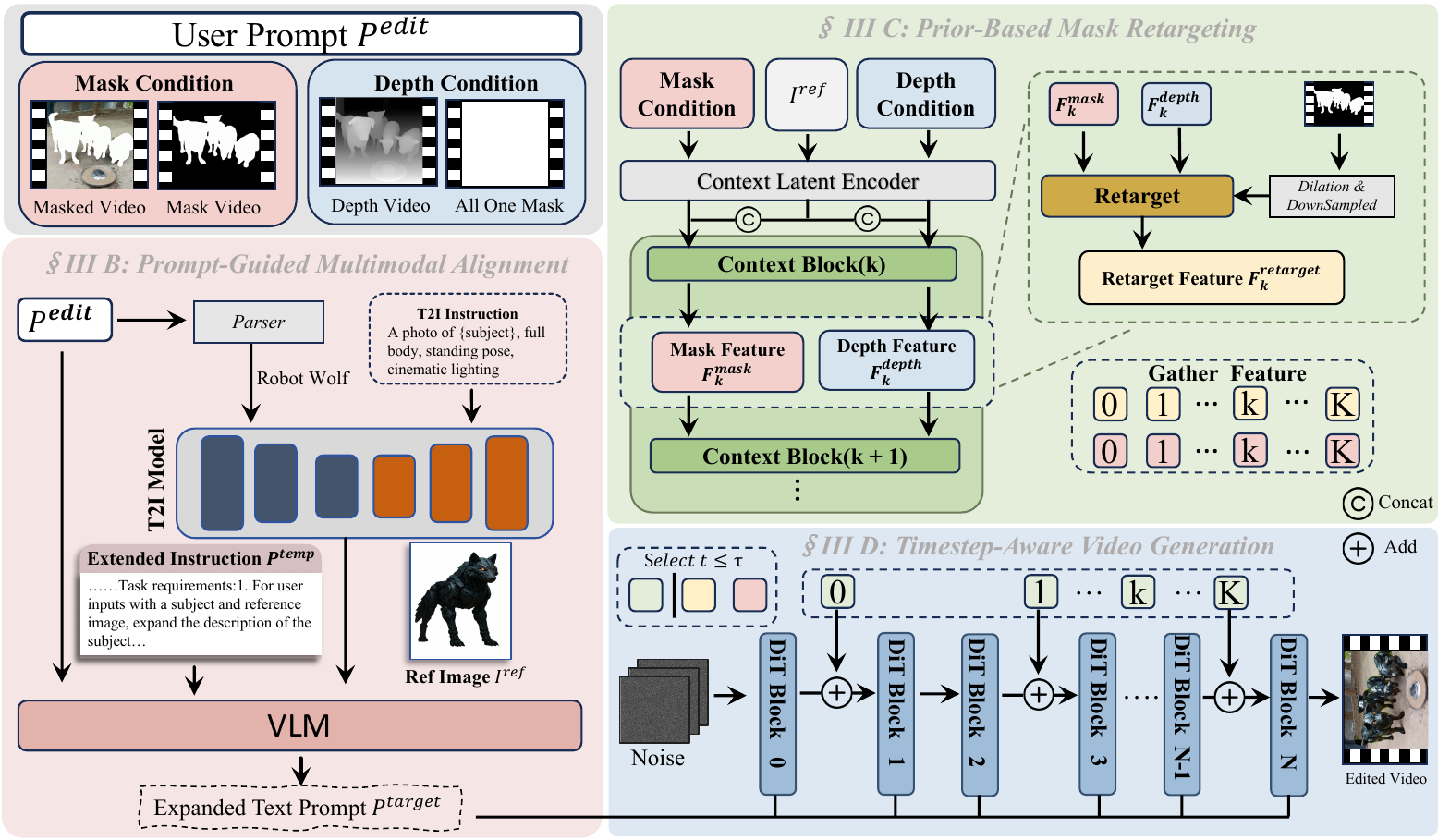} 
\end{center}
\vspace{-0.3cm}
\caption{\textbf{Overview of the ASTRA framework.} It first derives robust multimodal cues via a prompt-guided multimodal alignment. 
Then, a prior-based mask retargeting module produces a time-consistent mask sequence aligned with the input video. 
Finally, the multimodal cues and mask sequence are fed into a video generation model to synthesize the edited video.}
\label{fig:pipeline} 
\vspace{-0.3cm}
\end{figure*}

\section{Related Work}~\label{sec:rw}

\noindent\textbf{Video Editing.}
Early video editing methods primarily relied on GANs~\cite{gan,mittal2017sync,pan2017create,li2018video} via warping and rendering pipelines. Recently, diffusion models~\cite{rombach2022high,peebles2023scalable,ruiz2023dreambooth} have significantly advanced video synthesis. Many approaches fine tune text to image models~\cite{wu2023tune,qi2023fatezero,liu2024video,zhang2025r} on single videos for stylization and subject replacement. However, these one shot paradigms frequently overfit and struggle to generalize in dense multi-subject scenes. Alternatively, mask guided methods~\cite{wang2025videodirector,yang2025videograin,jiang2025vace,bian2025videopainter} improve spatial localization but depend heavily on initial mask quality. In complex scenarios, overlapping instances cause severe mask boundary entanglement. Consequently, the absence of an integrated mechanism to reconcile competing regional semantics often results in attribute leakage and fragmented transformations during simultaneous multi-subject editing.

\noindent\textbf{Instance Segmentation.}
Instance segmentation has evolved from early iterative region proposal refinement~\cite{rother2004grabcut} to modern deep learning approaches employing direct mask regression via cascade networks~\cite{ren2016faster,he2017mask,li2017fully} and indirect regression using heatmaps or queries~\cite{zhang2021k,cheng2022masked}.
Recently, prompted segmentation foundation models~\cite{kirillov2023segany,ravi2024sam,ren2024grounded} have demonstrated strong cross domain generalization. Nevertheless, these general purpose models still struggle in dense multi-subject environments, where closely interacting instances lead to boundary entanglement and mask leakage. Furthermore, the lack of explicit spatial temporal coordination between individual masks often results in identity drift and structural artifacts during temporal propagation, significantly hindering high fidelity video editing in complex scenarios.

\noindent\textbf{Text to Video Generation.}
Image to video (I2V) generation~\cite{singer2022make,yang2024cogvideox} has rapidly advanced video synthesis. Early diffusion based works~\cite{guo2023animatediff} inserted temporal layers into pretrained 2D U Nets~\cite{ronneberger2015u} for static to dynamic conversion. For instance, VideoPainter~\cite{bian2025videopainter} uses a dual branch design with diffusion transformers for mask guided inpainting. Concurrently, specialized I2V frameworks~\cite{singer2022make,ho2022imagen} trained on large scale datasets achieved remarkable results. Among them, DiT based models~\cite{hong2022cogvideo,yang2024cogvideox,wan2025wan,gao2025wan} offer superior global coherence and fine grained control. Leveraging these advantages, ASTRA adopts Wan2.1~\cite{wan2025wan} as its mask-driven I2V backbone to harness strong generative priors for high fidelity multi-subject editing. By integrating these robust priors with our specific alignment mechanisms, ASTRA effectively reconciles competing semantic signals to ensure strict spatial temporal coherence even in the most challenging multi-subject scenarios.


\section{Method}~\label{sec:method}
The ASTRA framework, illustrated in Fig.~\ref{fig:pipeline}, is designed to overcome the boundary entanglement and attention dilution typical of dense multi-subject scenes. Recognizing that general purpose models often struggle with complex overlapping instances, ASTRA introduces a synergistic pipeline to ensure temporal stability and prevent attribute leakage. We first review diffusion transformer preliminaries in Sec.~\ref{subsec:dit}, then detail our core contributions: prompt guided multimodal alignment (Sec.~\ref{subsec:alignment}), prior based mask retargeting (Sec.~\ref{subsec:retargeting}), and timestep aware video generation (Sec.~\ref{subsec:generation}).

\subsection{Preliminaries}\label{subsec:dit}
To incorporate structural controls without altering the primary diffusion model, we utilize an independent Context Module \cite{jiang2025vace} to process external spatiotemporal conditions and achieve precise feature alignment with the main generation pathway. Following the condition injection mechanism introduced in VACE \cite{jiang2025vace}, the Context Module replicates the denoising diffusion transformer (DiT) blocks of the pre-trained backbone (e.g., Wan2.1~\cite{wan2025wan}) to retain its robust spatiotemporal priors. During optimization, the primary generation model remains frozen. 
To balance computational efficiency and alignment precision, the module contains $K = N/2$ blocks, where $N$ is the total block count of the main network. In our framework, the Context Module primarily processes mask and depth condition groups. The mask input is formulated by concatenating the masked video with its binary mask along the channel dimension, while the depth sequence is concatenated with an all-ones mask. To integrate these signals smoothly, we adopt a skip-block alignment strategy. At each Context Module block $k \in \{1, \dots, K\}$, the extracted multi-scale condition features, denoted as $F_{k}^{mask}$ and $F_{k}^{depth}$, are directly injected into the corresponding $2k$-th block of the primary generation model.

\subsection{Prompt-Guided Multimodal Alignment}\label{subsec:alignment}
\begin{figure}[t]
\centering
\includegraphics[width=0.95\linewidth]{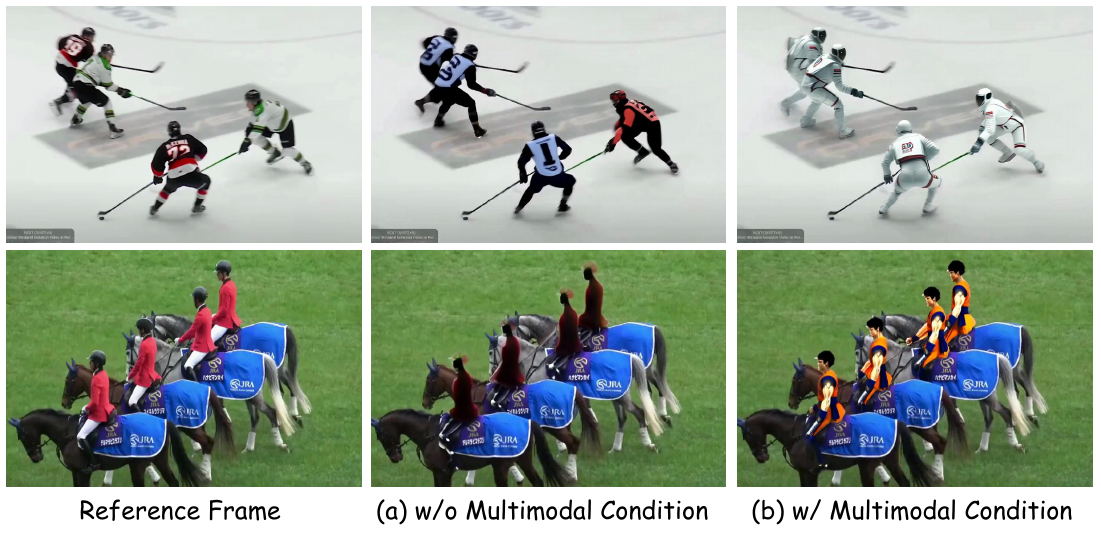}
\vspace{-0.2cm}
\caption{\textbf{Ablation of multimodal conditioning.} Top row: Hockey Players $\rightarrow$ Astronauts; Bottom row: Horse Riders $\rightarrow$ Gokus. Multimodal conditions ensure strict prompt adherence, enabling complete semantic transformations and mitigating attribute leakage across multiple subjects.}
\vspace{-0.5cm}
\label{fig:pgma_results}
\end{figure}

Recent studies~\cite{yin2023dragnuwa,singer2022make} show that the limited understanding ability of text encoders in video editing models often causes inconsistencies between editing results and intended semantics when using naive text prompts. In multi-subject editing scenarios, this issue becomes more pronounced. As shown in Fig.~\ref{fig:pgma_results} (a) top row, neighboring subjects dilute attention, and a naive prompt fails to impose a clear constraint on the "astronaut", thus not triggering the intended edit. Another case is shown in the bottom row of Fig.~\ref{fig:pgma_results} (a), where insufficient textual semantic constraints cause a semantic mismatch, making the attributes related to "Goku" only partially affect the target. These observations indicate that multi-subject settings require stronger multimodal alignment and subject-level control to ensure precise binding of editing intent and temporal stability. 

Based on these observations, we introduce a prompt-guided multimodal alignment module to explicitly realize cross-modal alignment and produce stable multimodal conditions. Specifically, as shown in Fig.~\ref{fig:pipeline}, we extract subject-specific tokens from the original editing prompt $P_{\text{edit}}$ to query a pretrained text-to-image model~\cite{podell2023sdxl}, generating a visual prior $I_{\text{ref}}$. This image bridges the abstract textual description with a concrete visual instance, anchoring the subject’s appearance. We then feed $I_{\text{ref}}$ and $P_{\text{edit}}$ into a vision-language model (VLM)~\cite{achiam2023gpt,wang2024qwen2}. Using an extended instruction template $P_{\text{temp}}$, the VLM interprets the visual attributes in $I_{\text{ref}}$ and expands the description in $P_{\text{edit}}$ in a controlled manner. This yields an enriched and visually grounded textual condition $P_{\text{target}}$:
\begin{equation}
P_{\text{target}} \;=\; \Phi_{\text{VLM}}\!\big(P_{\text{edit}},\, I_{\text{ref}} \,\big|\, P_{\text{temp}}\big),
\end{equation}
where $\Phi_{\text{VLM}}$ denotes the VLM function that reconciles the semantic intent in $P_{\text{edit}}$ with the structure and appearance priors provided by $I_{\text{ref}}$. Consequently, the complete multimodal signal comprises both the expanded text $P_{\text{target}}$ and the reference image $I_{\text{ref}}$. To effectively integrate these aligned conditions into the generation process, $I_{\text{ref}}$ is prepended as the first frame of the input sequence to serve as a direct visual anchor, while the enriched textual prompt $P_{\text{target}}$ is injected into the ViT blocks via cross-attention mechanisms. As shown in Fig.~\ref{fig:pgma_results} (b), grounding the editing process in this explicit multimodal alignment improves the fidelity of subject attributes and mitigates attention dilution and semantic drift, resulting in more coherent and targeted video edits.

\subsection{Prior-Based Mask Retargeting}\label{subsec:retargeting}
The accuracy of instance masks directly dictates the controllability and temporal stability of mask-driven video editing. In dense, multi-subject scenes, general-purpose segmentation models often fail to produce precise boundaries or capture hierarchical occlusion orders, leading to mask leakage and temporal inconsistencies. To overcome this, we introduce the prior-based mask retargeting module, as shown in Fig.~\ref{fig:pipeline}. Constrained by depth priors, this module spatially re-estimates instance boundaries and generates a temporally consistent retargeted mask sequence, significantly reducing leakage.

\begin{figure}[t]
\centering
\includegraphics[width=0.9\linewidth]{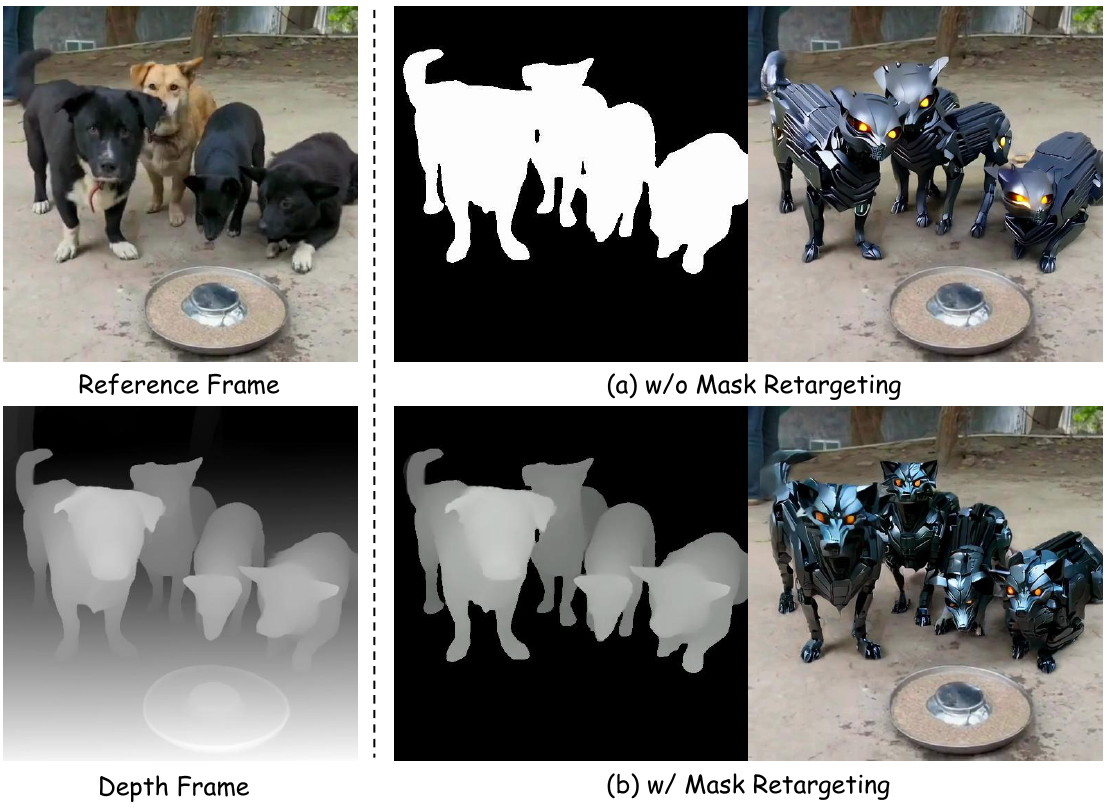}
\vspace{-0.2cm}
\caption{\textbf{Ablation of mask retargeting (Dogs $\rightarrow$ Robot Wolves).} Mask retargeting ensures strict spatial adherence to subjects across frames, yielding crisp boundaries and significantly mitigating mask leakage in dense scenes.}
\label{fig:pbmr}
\vspace{-0.3cm}
\end{figure}

Given an original video sequence $V_{ori}=\{v_{1},v_{2},...,v_{N}\}$ with $N$ pixel-level frames, we obtain the initial binary instance masks $M=\{m_{1},m_{2},...,m_{N}\}${~\cite{ren2024grounded}} and the corresponding depth maps $D=\{d_{1},d_{2},...,d_{N}\}${~\cite{yang2024depthv2}}. To compute the mask-guided condition, we first obtain the masked video via element-wise multiplication strictly using the hard binary mask: $V_{masked}=V_{ori}\odot M$. These pixel-space inputs are then encoded into the latent space, resulting in sequences of $n$ latent frames (where $n \le N$ due to temporal compression). Importantly, the Context Module maintains its original forward pass without any architectural modifications. During this standard forward pass across $K$ context blocks, it extracts multi-scale condition features. Specifically, let $F_{i,k}^{mask}$ and $F_{i,k}^{depth}$ denote the latent mask feature and latent depth feature, respectively, for the $i$-th frame at the $k$-th block.

To ensure a smooth transition and precise spatial alignment between these modality features, we apply morphological dilation to expand the initial editing region. For each binary mask $m_{i}$, the dilated mask is defined as:
\begin{equation}
m_{i}^{\prime}[p,q]=\max_{(u,v)\in\mathcal{N}_{c}}m_{i}[p+u,q+v]
\end{equation}
where $\mathcal{N}_{c}$ is a square neighborhood of size $c\times c$. A Gaussian filter is then applied to $m_{i}^{\prime}$, and the result is spatially and temporally downsampled to match the latent resolution, producing a soft mask sequence $\tilde{M}=\{\tilde{m}_{1},\tilde{m}_{2},...,\tilde{m}_{n}\}$.

At each block $k \in \{1, \dots, K\}$ and for each frame $i \in \{1, \dots, n\}$, we achieve precise region modification by explicitly aligning the condition features. The layer-wise retargeted feature $F_{i,k}^{retarget}$ at each spatial location $(x,y)$ is formulated as:
\begin{equation}
\begin{split}
F_{i,k}^{\text{retarget}}(x,y) 
    &= \tilde{m}_i(x, y) F_{i,k}^{\text{depth}}(x, y) \\
    &\quad + (1 - \tilde{m}_i(x, y)) F_{i,k}^{\text{mask}}(x, y).
\end{split}
\end{equation}
This alignment strategy guarantees that within the target subjects (high $\tilde{m}_{i}$), the generative process is primarily guided by depth to recover proper geometry and occlusion ordering, whereas in the background (low $\tilde{m}_{i}$), mask constraints dominate. Finally, we gather these aligned features across all $K$ layers to form the comprehensive retarget feature set $F^{retarget}=\{F_{k}^{retarget}\}_{k=1}^{K}$, which is subsequently injected into the primary video generation model.

\begin{figure}[t]
\centering
\includegraphics[width=0.95\linewidth]{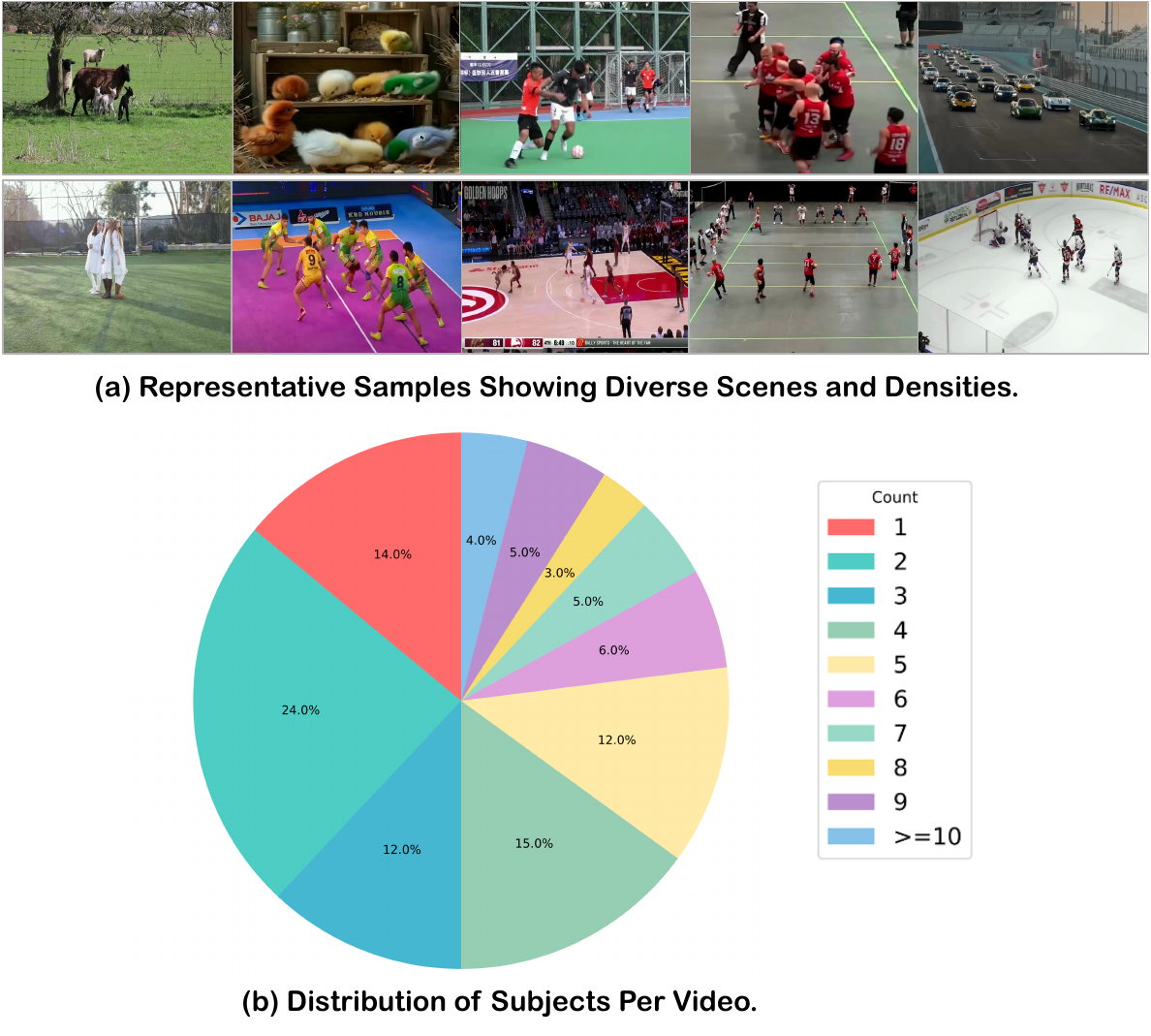}
\vspace{-0.3cm}
\caption{\textbf{Overview of MSVBench.} (a) Representative samples across diverse scenes and subject densities. (b) Statistical distribution of subject counts per video.}
\label{fig:data_num}
\vspace{-0.5cm}
\end{figure}

\subsection{Timestep-Aware Video Generation}\label{subsec:generation}
Given the comprehensive retarget feature set $F^{retarget}$ generated by the Context Module, we can explicitly condition various mask-driven video generator by integrating these features into the main ViT backbone \cite{wan2025wan,gao2025wan,zhang2023adding,jiang2025vace}. Although $F^{retarget}$ can in principle be injected at all denoising steps, continuing this structural alignment at late stages produces severe artifacts and unnatural seams. This occurs because early denoising steps primarily shape low-frequency structures, whereas late steps focus on refining high-frequency visual details \cite{wu2024freediff,qian2024boosting}.
We therefore adopt a timestep-aware injection strategy: the retargeted features are utilized exclusively during the early structural phase, and the model reverts to mask-only conditioning for the subsequent refinement phase. Let $T$ be the total number of denoising steps and $\tau$ be the injection threshold. At a given denoising step $t \in \{1, \dots, T\}$, the final conditional feature $F^{cond}_{t}$ is formulated as:
\begin{equation}
F^{cond}_{t}=
\begin{cases}
F^{retarget}_{t}, & t\le \tau,\\[2pt] 
F^{mask}_{t}, & t> \tau.
\end{cases}
\label{eq:condition_injection}
\end{equation}
This phased alignment scheme accurately tracks the motion and hierarchical occlusion encoded by the retarget sequence during structural formation, whilst preserving high-quality textural details during final refinement. Furthermore, it generalizes seamlessly across different diffusion architectures, yielding consistent visual gains in dense, multi-subject scenarios.

\begin{table}[t] 
\centering
\vspace{-0.3cm}
\caption{\textbf{Quantitative comparison on MSVBench.} ASTRA is benchmarked against diverse state of the art baselines. The best and second best scores are \textbf{bold} and \underline{underlined}, respectively. Gray text denotes closed source models.}
\label{tab:sota}
\renewcommand{\arraystretch}{1.0}
\setlength{\tabcolsep}{3pt} 
\resizebox{\linewidth}{!}{%
\begin{tabular}{l|ccccc}
\toprule
\rowcolor{mygray1}
\multicolumn{1}{l|}{Methods} &
\multicolumn{1}{c}{Warp-Err ($\downarrow$)} &
\multicolumn{1}{c}{CLIP-T ($\uparrow$)} &
\multicolumn{1}{c}{CLIP-F ($\uparrow$)} &
\multicolumn{1}{c}{Q-Edit ($\uparrow$)} &
\multicolumn{1}{c}{CM-Err ($\downarrow$)} \\
\midrule
FateZero \cite{qi2023fatezero}    & 2.16 & 24.26 & 97.42 & 11.23 & 3.49 \\
TokenFlow \cite{geyer2023tokenflow}   & 2.10 & \underline{26.57} & 96.93 & 12.65 & 3.78 \\
VideoPainter\cite{bian2025videopainter} & 2.05 & 24.13 & 96.97 & 11.77 & 4.70 \\
VideoGrain  \cite{yang2025videograin} & 1.98 & 24.71 & 97.13 & 12.48 & 3.12 \\
DMT        \cite{yatim2024space}   & \underline{1.87} & 24.55 & 96.76 & 13.13 & 3.80 \\
VACE       \cite{jiang2025vace}   & \underline{1.87} & 24.78 & \textbf{97.94} & \underline{13.24} & \underline{3.00} \\
\midrule
\textcolor{mygray}{Kling}  & \textcolor{mygray}{2.00} & \textcolor{mygray}{25.66} & \textcolor{mygray}{98.37} & \textcolor{mygray}{12.83} & \textcolor{mygray}{4.39} \\
\textcolor{mygray}{Runway} & \textcolor{mygray}{1.87} & \textcolor{mygray}{25.82} & \textcolor{mygray}{97.73} & \textcolor{mygray}{13.81} & \textcolor{mygray}{3.36} \\
\textcolor{mygray}{Viggle} & \textcolor{mygray}{1.86} & \textcolor{mygray}{25.04} & \textcolor{mygray}{97.53} & \textcolor{mygray}{13.46} & \textcolor{mygray}{3.43} \\
\midrule
ASTRA & \textbf{1.85} & \textbf{27.23} & \underline{97.93} & \textbf{14.72} & \textbf{2.83} \\
\bottomrule
\end{tabular}}
\vspace{-0.5cm}
\end{table}

\section{Experiments}~\label{sec:exp}
\subsection{Experimental Setup}
\noindent\textbf{Datasets.}
To evaluate multi-subject video editing in complex scenarios, we construct MSVBench. It comprises 100 videos sourced from YouTube and TikTok, covering diverse categories such as humans, animals, and vehicles. The benchmark explicitly emphasizes dense layouts, strong occlusions, and dynamic camera motions. Over 60\% of the videos contain three or more subjects, ranging up to ten (Fig.~\ref{fig:data_num}). We utilize GPT-4o~\cite{achiam2023gpt} to generate verified scene descriptions and editing prompts. Target instance masks are extracted using Grounded SAM2~\cite{ren2024grounded} and manually inspected for temporal consistency. The complete dataset, including videos, masks, and prompts, will be publicly released. Furthermore, to rigorously assess the cross-domain generalization capability of ASTRA, we extend our evaluation to the standard LOVEU-TGVE-2023~\cite{wu2023cvpr} benchmark. This dual-dataset evaluation strategy comprehensively validates both multi-subject handling and general editing fidelity.

\noindent\textbf{Evaluation Metrics.}
Following~\cite{cong2023flatten,yang2025videograin}, we employ four standard metrics: Warp-Err~\cite{teed2020raft} for background preservation, CLIP-T~\cite{radford2021learningtransferablevisualmodels} for text-region alignment, CLIP-F~\cite{radford2021learningtransferablevisualmodels} for temporal coherence, and Q-Edit~\cite{cong2023flatten} for overall editing fidelity. Additionally, we propose the Center Matching Error (CM-Err) to assess subject count and spatial layout consistency, as traditional pixel overlap metrics fail to capture topological changes like additions or merges. For frame $t$ with width $W$ and height $H$, let $\mathcal{A}_t=\{a_j\}$ and $\mathcal{B}_t=\{b_k\}$ be the bounding boxes detected by GroundingDINO~\cite{liu2024grounding} in the original and edited frames. The normalized center distance between a pair $(a_j,b_k)$ is:
\begin{equation}
d_{jk}^{(t)} \;=\; \frac{\big\|c(a_j)-c(b_k)\big\|_2}{\sqrt{W^2+H^2}} \;\in[0,1].
\end{equation}
Using $d_{jk}^{(t)}$ as the matching cost, we compute a minimal one-to-one assignment. Let $M_t$ be the number of matched pairs and $U_t=|\mathcal{A}_t|+|\mathcal{B}_t|-2M_t$ be the count of unmatched boxes. The frame-level error is formulated as:
\begin{equation}
\mathrm{CM\text{-}Err}^{(t)} \;=\; \frac{\sum_{i=1}^{M_t} d_i^{(t)} + U_t}{M_t + U_t}.
\end{equation}
The final CM-Err is the average error over $T$ frames. Lower values indicate superior layout and count preservation.

\noindent\textbf{Implementation Details.}
All experiments are conducted on a single NVIDIA A800 80 GB GPU. The denoising and conditional DiT models are initialized from Wan2.1~\cite{wan2025wan}. Multimodal priors are obtained using SDXL~\cite{podell2023sdxl} and Qwen2.5-VL~\cite{bai2025qwen25vltechnicalreport}. Video inputs are pre-processed with Grounded SAM 2~\cite{ren2024grounded} for masks and Depth Anything V2~\cite{yang2024depthv2} for depth maps. During inference, we perform 50 denoising steps and set the condition injection threshold to $\tau=30$.

\begin{table}[t]
\centering
\vspace{-0.3cm}
\caption{\textbf{Generalization results on LOVEU TGVE 2023.} Performance comparison against video editing baselines.}
\label{tab:sota2}
\renewcommand{\arraystretch}{1.0}
\setlength{\tabcolsep}{3pt}
\resizebox{\linewidth}{!}{%
\begin{tabular}{l|ccccc}
\toprule
\rowcolor{mygray1}
\multicolumn{1}{l|}{Methods} &
\multicolumn{1}{c}{Warp-Err ($\downarrow$)} &
\multicolumn{1}{c}{CLIP-T ($\uparrow$)} &
\multicolumn{1}{c}{CLIP-F ($\uparrow$)} &
\multicolumn{1}{c}{Q-Edit ($\uparrow$)} &
\multicolumn{1}{c}{CM-Err ($\downarrow$)} \\
\midrule
VideoGrain~\cite{yang2025videograin}   & 2.14 & \underline{25.22} & 96.78 & 11.78 & \underline{3.05} \\
TokenFlow~\cite{geyer2023tokenflow}   & 2.23 & 24.22 & \underline{97.20} & 10.86 & 3.54 \\
FateZero~\cite{qi2023fatezero}        & 2.18 & 23.80 & 97.11 & 10.91 & 3.75 \\
DMT~\cite{yatim2024space}             & \textbf{1.90} & 23.82 & 97.18 & \underline{12.53} & 3.61 \\
VideoPainter~\cite{bian2025videopainter} & 2.12 & 22.95 & 95.95 & 10.82 & 4.04 \\
\midrule
ASTRA                      & \underline{2.04} & \textbf{25.99} & \textbf{97.23} & \textbf{12.74} & \textbf{2.66} \\
\bottomrule
\end{tabular}}
\vspace{-0.5cm}
\end{table}

\begin{figure*}[t]
\begin{center}
\includegraphics[width=0.95\linewidth]{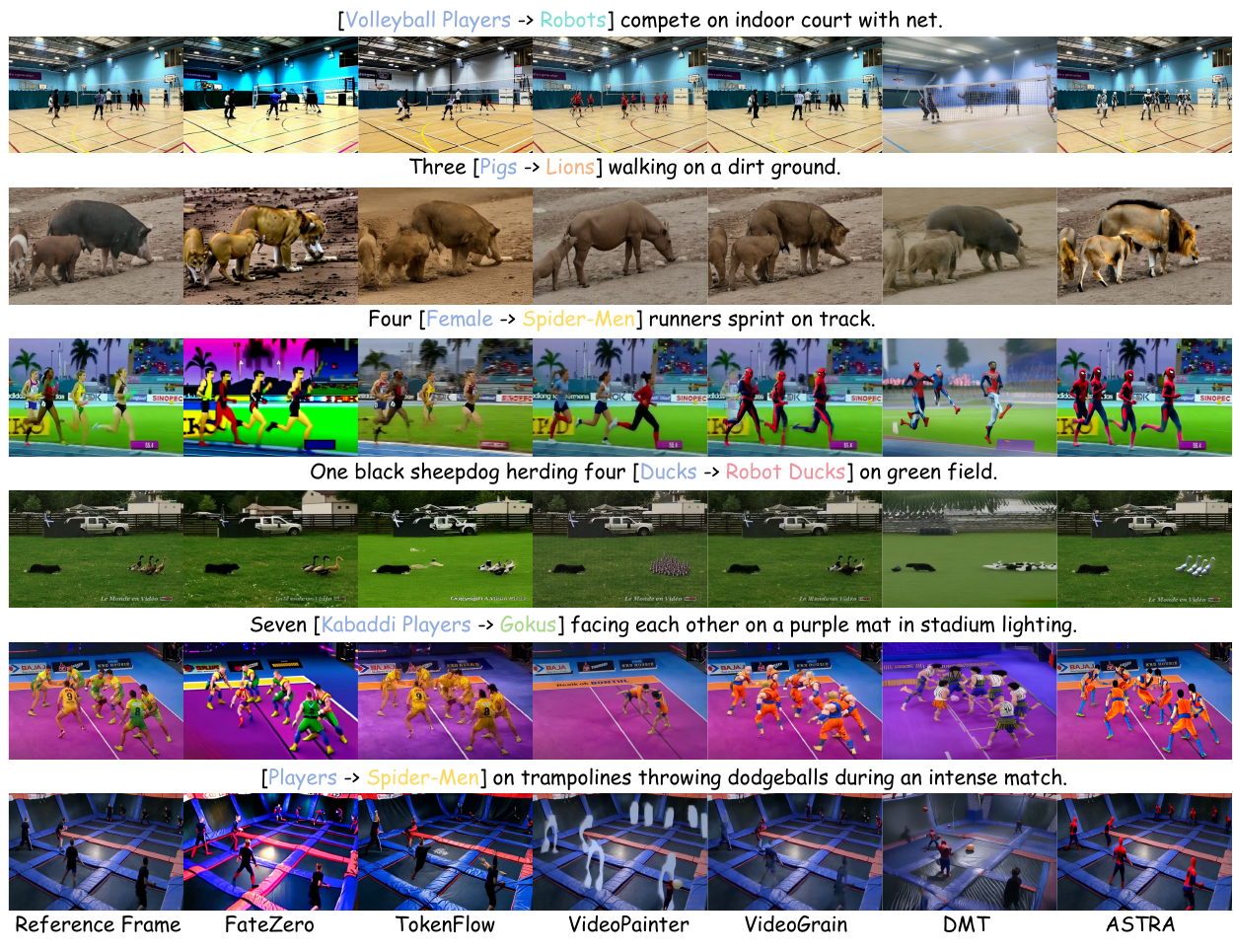}
\end{center}
\vspace{-0.5cm}
\caption{\textbf{Qualitative comparison on MSVBench.} ASTRA is benchmarked against diverse state of the art baselines across complex scenarios involving multiple subjects. More videos are available at \url{https://muzishen.github.io/ASTRA/}.}
\label{fig:sota} 
\vspace{-0.5cm}
\end{figure*}

\subsection{Comparison with State-of-the-Art}
We benchmark ASTRA against a diverse set of state-of-the-art video editing methods. Our baselines include prominent open-source frameworks, specifically FateZero~\cite{qi2023fatezero}, TokenFlow~\cite{geyer2023tokenflow}, VideoPainter~\cite{bian2025videopainter}, VideoGrain~\cite{yang2025videograin}, and DMT~\cite{yatim2024space}. Additionally, we extend our evaluation to leading closed-source commercial systems, including Kling\footnote{\url{https://klingai.com/global}}, Runway\footnote{\url{https://runwayml.com}}, and Viggle\footnote{\url{https://viggle.ai}}.

\noindent\textbf{Quantitative Results.} 
From TABLE~\ref{tab:sota}, ASTRA achieves state-of-the-art performance on MSVBench. Specifically, it attains the best scores in Warp-Err (1.85), CLIP-T (27.23), Q-Edit (14.72), and CM-Err (2.83), with a highly competitive CLIP-F (97.93). Compared to the strongest open-source baseline (VACE\cite{jiang2025vace}), ASTRA improves Q-Edit by 11.2\% (from 13.24 to 14.72), proving that our prompt-guided multimodal alignment significantly enhances edit fidelity. It also reduces CM-Err to 2.83 and Warp-Err to 1.85, demonstrating that our prior-based mask retargeting effectively preserves spatial layouts and background stability. Notably, ASTRA even outperforms closed-source commercial models like Runway in Q-Edit and CM-Err.
To validate cross-domain generalization, we evaluate ASTRA on LOVEU-TGVE-2023\cite{wu2023cvpr}, where over 80\% of videos feature single or paired subjects. From TABLE~\ref{tab:sota2}, ASTRA consistently delivers the highest semantic consistency and editing quality (CLIP-T 25.99, CLIP-F 97.23, Q-Edit 12.74). It also records the lowest CM-Err (2.66), indicating superior layout preservation. While yielding a competitive Warp-Err of 2.04 (surpassed only by DMT\cite{yatim2024space}), ASTRA exhibits the most balanced performance profile. These results confirm that our method not only excels in dense scenes but also generalizes robustly to conventional few-subject video editing tasks.

\noindent\textbf{Qualitative Results.}
As illustrated in Fig.~\ref{fig:sota}, competing methods like FateZero\cite{qi2023fatezero}, TokenFlow\cite{geyer2023tokenflow}, and VideoGrain\cite{yang2025videograin} suffer from severe boundary entanglement and attention dilution. This results in incomplete edits, background corruption, and attribute leakage across interacting subjects. In contrast, ASTRA accurately transforms all designated targets while strictly preserving non-target regions, proving that our prompt-guided multimodal alignment provides robust conditions for precise subject conversion. Furthermore, under complex limb motions and heavy occlusions, approaches such as VideoPainter\cite{bian2025videopainter} and DMT\cite{yatim2024space} frequently exhibit missing subjects or degraded visual fidelity. Conversely, our prior-based mask retargeting ensures temporally coherent mask sequences. This allows ASTRA to maintain strict frame-to-frame consistency and high structural fidelity even amidst complex scene dynamics.

\begin{figure}[t]
\centering
\includegraphics[width=0.9\linewidth]{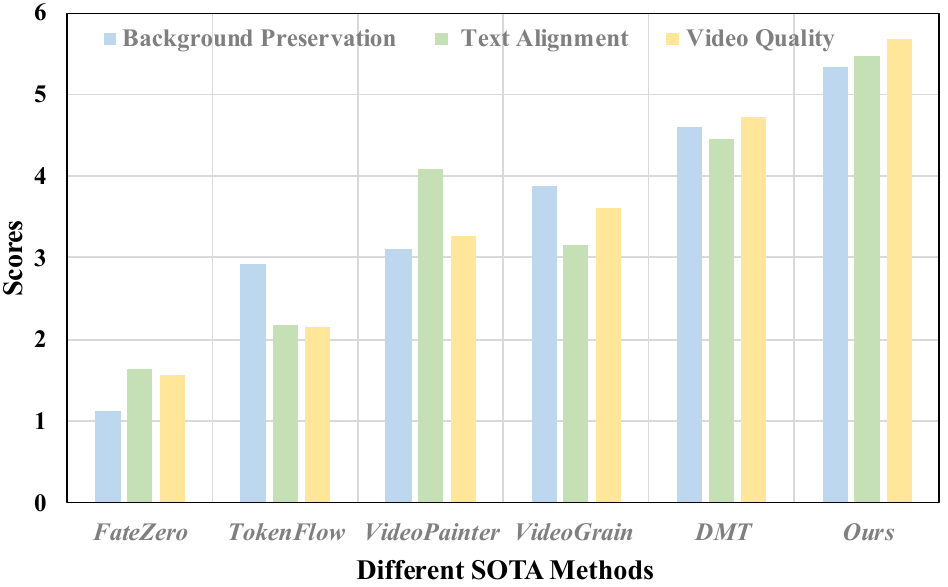}
\vspace{-0.1cm}
\caption{\textbf{Human evaluation results.} Higher scores across the three metrics indicate superior performance.}
\label{fig:user_study}
\vspace{-0.6cm}
\end{figure}

\noindent\textbf{Human Evaluation.}
To complement our objective metrics with human perception, we conducted a user study involving 20 volunteers and 20 randomly selected videos. Participants ranked the editing outcomes of all competing methods based on three critical criteria: Background Preservation (BP), Text Alignment (TA), and Video Quality (VQ). As illustrated in Fig.~\ref{fig:user_study}, ASTRA consistently secured the highest rankings across all three dimensions. These subjective evaluations strongly corroborate our quantitative findings, confirming that ASTRA delivers superior visual fidelity and robust performance in complex multi-subject scenarios.

\begin{table}[t]
\centering
\vspace{-0.3cm}
\caption{\textbf{Quantitative ablation analysis.} Evaluation of individual framework components and their contributions to overall performance. The best score is \textbf{bold}.}
\label{tab:ablation}
\setlength{\tabcolsep}{6pt} 
\begin{tabular}{l|c|c|c}
\toprule[0.4mm]
\rowcolor{mygray1}
\cellcolor{mygray1}Methods & CLIP-T $\uparrow$ & Q-Edit $\uparrow$ & CM-Err $\downarrow$\\ \midrule
B0 & 24.78 & 13.24 & 3.00 \\
B1 & 25.10 & 13.42 & 2.87 \\
B2 & 26.12 & 14.04 & 2.99 \\
\textbf{ASTRA} & \textbf{27.23} {\color[RGB]{0,150,0}\scriptsize (+9.9\%)} & \textbf{14.72} {\color[RGB]{0,150,0}\scriptsize (+11.2\%)} & \textbf{2.83} {\color[RGB]{0,150,0}\scriptsize (-5.7\%)} \\ 
\bottomrule[0.4mm]
\end{tabular}
\vspace{-0.4cm}
\end{table}

\begin{figure}[t]
\centering
\includegraphics[width=\linewidth]{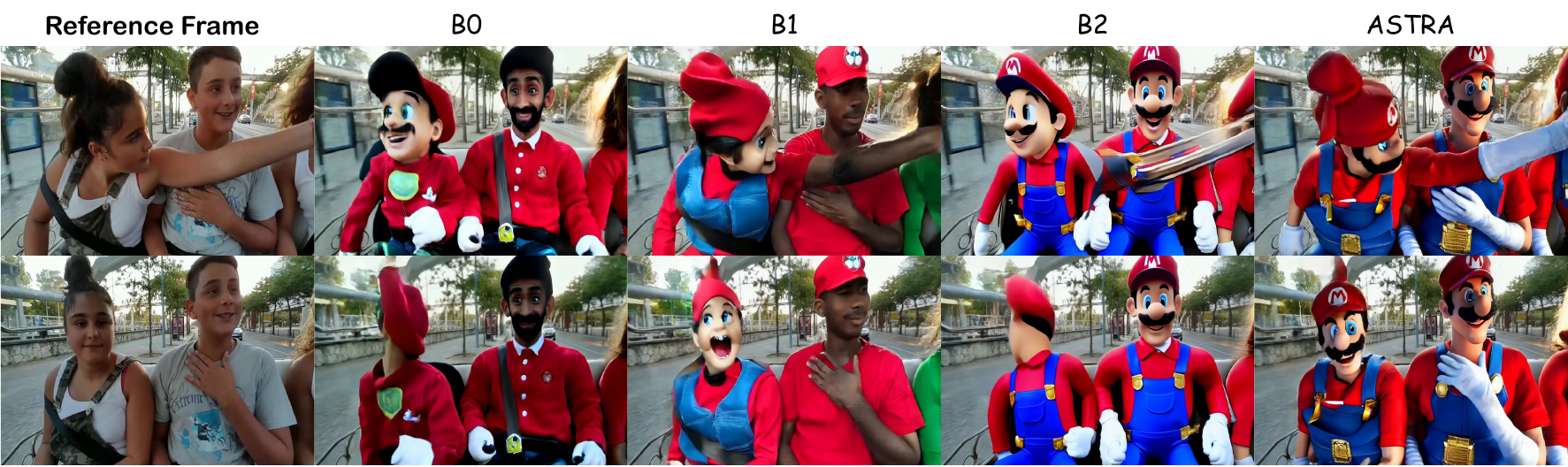}
\vspace{-0.3cm}
\caption{\textbf{Qualitative ablation of ASTRA (People $\rightarrow$ Super Mario).} This visualization demonstrates the contribution of individual modules toward achieving consistent and high quality subject transformations.}
\label{fig:ablation}
\vspace{-0.5cm}
\end{figure}

\subsection{Ablation Study}
\noindent\textbf{Effectiveness of Core Components.}\label{eff_module}
To evaluate the contribution of each component in ASTRA, we conduct ablations over three variants (B0–B2) against the full pipeline, keeping all other settings fixed:
\begin{itemize}
    \item \textbf{B0}: The base mask-driven video generation model (Wan2.1) without any additional modules.
    \item \textbf{B1}: Enables only the prior-based mask retargeting module, relying on the original text prompt.
    \item \textbf{B2}: Enables only the prompt-guided multimodal alignment module, using the original unretargeted mask sequence.
\end{itemize}
As shown in TABLE~\ref{tab:ablation}, integrating the prompt-guided multimodal alignment module (B2) substantially improves text alignment and overall fidelity over the base model (B0), increasing CLIP-T from 24.78 to 26.12 and Q-Edit from 13.24 to 14.04. This demonstrates that explicitly aligning textual prompts with visual priors provides robust multimodal conditioning. Visually, as depicted in Fig.~\ref{fig:ablation}, B2 effectively mitigates attribute leakage and incomplete edits, ensuring accurate transformations across multiple subjects (e.g., proper Super Mario attire).
Furthermore, incorporating the prior-based mask retargeting module (B1) notably enhances spatial consistency and layout preservation. B1 effectively reduces the CM-Err from 3.00 (B0) to 2.87, alongside steady gains in CLIP-T (25.10) and Q-Edit (13.42). Fig.~\ref{fig:ablation} illustrates that B1 successfully suppresses boundary entanglement, ensuring the generated content adheres strictly to the target boundaries without corrupting non-target regions.
Finally, the full ASTRA framework integrates both components, achieving the best performance across all metrics (e.g., peak Q-Edit of 14.72 and lowest CM-Err of 2.83). This confirms the synergistic effect of precise multimodal conditions and temporally consistent mask retargeting in complex multi-subject scenarios.

\begin{table}[t]
\centering
\vspace{-0.3cm}
\caption{\textbf{Ablation of conditioning modalities.} Performance comparison across text, image, and combined input conditions.}
\label{tab:ablation_results}
\renewcommand{\arraystretch}{1.0}
\setlength{\tabcolsep}{4pt}
\resizebox{\linewidth}{!}{%
\begin{tabular}{l|cccc}
\toprule
\rowcolor{mygray1}
\multicolumn{1}{l|}{Methods} & 
\multicolumn{1}{c}{CLIP-T ($\uparrow$)} & 
\multicolumn{1}{c}{CLIP-F ($\uparrow$)} & 
\multicolumn{1}{c}{Q-Edit ($\uparrow$)} & 
\multicolumn{1}{c}{CM-Err ($\downarrow$)} \\ 
\midrule
Text Only                          & 24.33 & 97.24 & 13.06 & 3.00 \\
Image Only                         & 24.41 & 97.59 & 13.26 & 3.38 \\
Text + Image                       & 25.45 & 97.80 & 13.68 & 3.01 \\
Ours & \textbf{27.23} & \textbf{97.93} & \textbf{14.72} & \textbf{2.83} \\
\bottomrule
\end{tabular}}
\vspace{-0.5cm}
\end{table}

\begin{figure}[t]
\begin{center}
\includegraphics[width=0.95\linewidth]{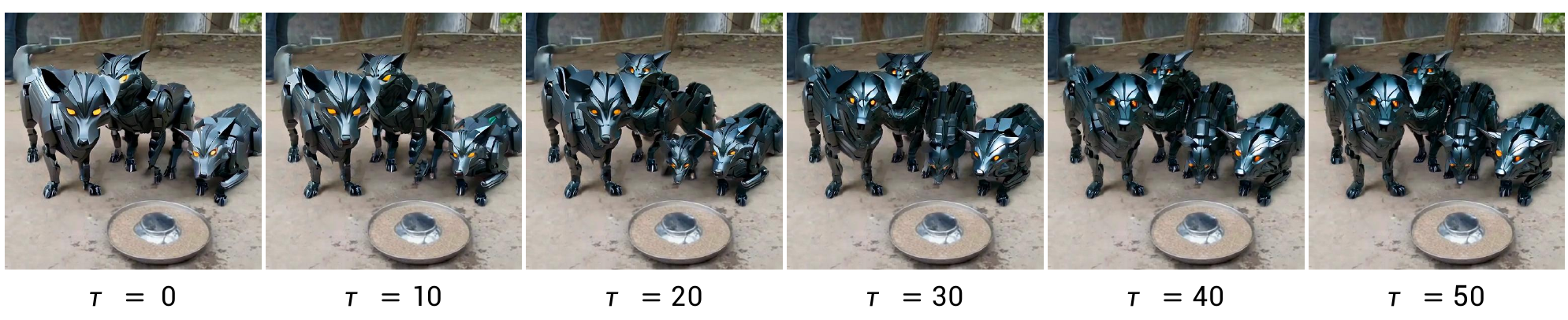}
\end{center}
\vspace{-0.3cm}
\caption{\textbf{Ablation of injection threshold $\tau$.} Variations in $\tau$ illustrate the trade off between structural constraints and high frequency detail synthesis. Visualizations represent the final frame of the edited video sequence.}
\label{fig:tao} 
\vspace{-0.5cm}
\end{figure}

\begin{table}[t]
\centering
\vspace{-0.3cm}
\caption{\textbf{Ablation of soft mask parameters.} Impact of varying dilation window size $c$ and Gaussian filter size $g$. Compared to the hard mask $(0, 0)$, soft spatial transitions consistently enhance semantic alignment and layout preservation.}
\label{tab:parameters-ablation}
\renewcommand{\arraystretch}{1.0}
\setlength{\tabcolsep}{4pt} 
\resizebox{\linewidth}{!}{%
\begin{tabular}{l|ccccc}
\toprule
\rowcolor{mygray1}
\multicolumn{1}{l|}{$(c, g)$} & 
\multicolumn{1}{c}{Warp-Err ($\downarrow$)} & 
\multicolumn{1}{c}{CLIP-T ($\uparrow$)} & 
\multicolumn{1}{c}{CLIP-F ($\uparrow$)} & 
\multicolumn{1}{c}{Q-Edit ($\uparrow$)} & 
\multicolumn{1}{c}{CM-Err ($\downarrow$)} \\ 
\midrule
$(0, 0)$  & 1.85 & 26.91 & 97.94 & 14.56 & 2.90 \\
$(10, 2)$ & \textbf{1.83} & 26.92 & \textbf{97.98} & 14.71 & 2.83 \\
$(10, 8)$ & 1.86 & 27.21 & 97.95 & 14.63 & \textbf{2.73} \\
$(5, 4)$  & 1.85 & 27.07 & 97.92 & 14.64 & 2.82 \\
$(15, 4)$ & 1.84 & 27.01 & \textbf{97.98} & 14.68 & 2.81 \\
$(10, 4)$ & 1.85 & \textbf{27.23} & 97.93 & \textbf{14.72} & 2.83 \\
\bottomrule
\end{tabular}}
\end{table}

\noindent\textbf{Impact of Conditioning Modalities.}
To isolate the effects of different input conditions, we evaluate isolated and naively combined modalities in TABLE~\ref{tab:ablation_results}. Relying strictly on a single modality (Text Only or Image Only) yields suboptimal text alignment and layout preservation. While directly providing both inputs (Text + Image) offers marginal metric improvements, it remains insufficient for complex multi-subject scenarios and frequently suffers from attention dilution. In contrast, our explicit alignment strategy (Ours) achieves peak performance across all metrics, notably boosting CLIP-T to 27.23 and Q-Edit to 14.72. These results confirm that explicitly reconciling abstract textual semantics with concrete visual priors effectively mitigates attribute leakage and significantly elevates overall editing consistency.

\begin{figure}[t]
\centering
\includegraphics[width=\linewidth]{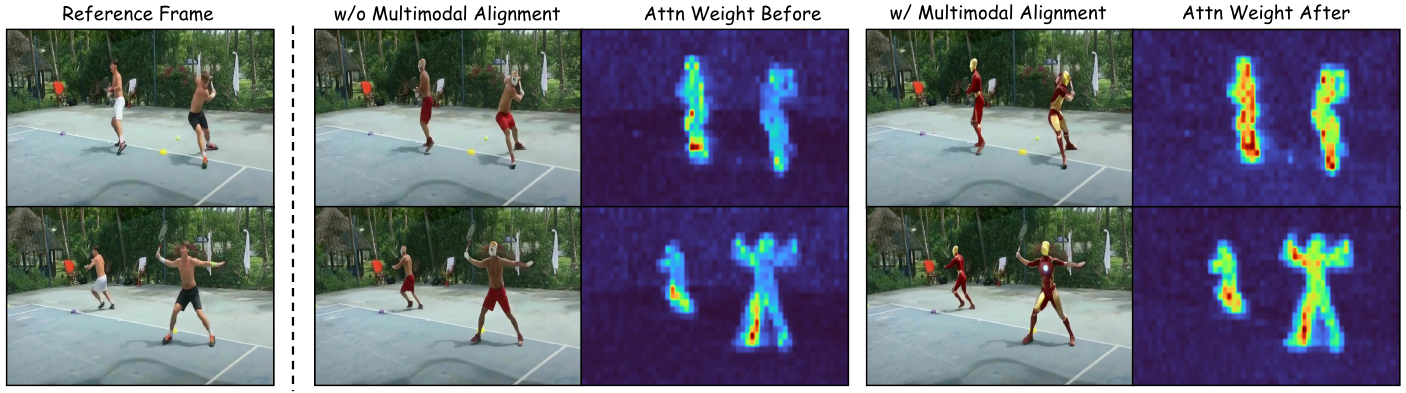}
\vspace{-0.3cm}
\caption{\textbf{Cross attention distribution.} Comparison of attention maps for target tokens without (w/o) and with (w/) multimodal conditioning. ASTRA ensures holistic semantic coverage across all edited subjects. (Players $\rightarrow$ Iron Men)}
\label{fig:attention}
\vspace{-0.5cm}
\end{figure}

\noindent\textbf{Injection Threshold.} 
We systematically vary the injection threshold $\tau$ from 0 to 50 to determine the optimal duration for structural guidance during the denoising process. As illustrated in Fig.~\ref{fig:tao}, excessively small values of $\tau$ provide insufficient structural constraints, inevitably leading to boundary leakage, imperfect occlusion ordering, and occasional identity drift. Conversely, excessively large values inappropriately extend fusion signals into the late refinement steps, introducing visual artifacts such as texture corruption and unnatural seams. An intermediate setting strikes the optimal balance. Specifically, near $\tau = 30$, the framework strictly preserves geometric structures and hierarchical layering while allowing the generative backbone to synthesize high frequency details, thereby producing clean boundaries and highly stable appearances. Consequently, we establish $\tau = 30$ as the default configuration based on cross validation over a held out split.

\noindent\textbf{Soft Mask Parameters.} 
We investigate the influence of the morphological dilation window size $c$ and the Gaussian filter size $g$ on soft mask generation. TABLE~\ref{tab:parameters-ablation} summarizes the impact of various parameter combinations. The performance remains notably stable across most configurations, demonstrating that ASTRA is highly robust to specific parameter variations. More importantly, compared to the hard mask baseline $(0,0)$, applying soft spatial transitions consistently reduces the CM-Err metric and enhances text alignment alongside overall editing quality. This confirms our initial design intuition: introducing soft masks effectively mitigates abrupt structural transitions between edited and unedited regions, thereby significantly improving layout preservation and temporal stability.

\subsection{Further Analysis and Extensions}
\noindent\textbf{Cross Attention Visualization.} 
To thoroughly investigate the underlying mechanism of our proposed method, we visualize the spatial distribution of cross attention weights in Fig.~\ref{fig:attention}. Given the target prompt ``Two Iron Men are playing tennis on a tennis court,'' we explicitly extract and analyze the attention maps corresponding to the target token ``Iron Men.'' Without the prompt guided multimodal alignment module, the attention weights remain notably sparse and inevitably collapse into highly localized regions, such as merely focusing on the subjects' heads. This severe attention dilution directly causes incomplete semantic transformations and subsequent attribute leakage. Conversely, by incorporating explicit multimodal conditions, ASTRA successfully achieves a dense, uniform, and accurate attention distribution across the entire bodies of both interacting subjects. This holistic feature binding ensures that the generative model comprehensively captures the designated target regions, ultimately resulting in complete, temporally stable, and visually coherent subject edits.

\noindent\textbf{Comparison with Iterative Editing Paradigms.}
We compare ASTRA against iterative, sequential editing using VACE~\cite{jiang2025vace} and VideoPainter~\cite{bian2025videopainter}. As TABLE~\ref{tab:results} shows, ASTRA excels in both generation quality and efficiency. Compared to VACE, ASTRA achieves superior semantic alignment (CLIP-T 27.23 vs. 25.24) and overall fidelity (Q-Edit 14.72 vs. 14.25). Crucially, repeated encoding cycles in iterative methods accumulate structural artifacts, leading to degraded layout preservation and significantly higher CM-Err scores (3.30 for VACE compared to 2.83 for ASTRA).
Furthermore, sequential processing drastically inflates computational costs. Iterative methods require over 18 minutes of inference time, whereas ASTRA completes complex multi-subject synthesis in nearly half the time (10m 08s). By concurrently resolving all targets in a single forward pass, ASTRA eliminates compounding errors and ensures highly efficient video editing.

\begin{table}[t]
\centering
\vspace{-0.3cm}
\caption{\textbf{Comparison with iterative state of the art baselines.} Despite operating as a single pass framework, ASTRA consistently outperforms iterative counterparts in both visual quality and inference efficiency.}
\label{tab:results}
\renewcommand{\arraystretch}{1.0}
\setlength{\tabcolsep}{4pt}
\resizebox{\linewidth}{!}{%
\begin{tabular}{l|cccc|c}
\toprule
\rowcolor{mygray1}
\multicolumn{1}{l|}{Methods} & 
\multicolumn{1}{c}{CLIP-T ($\uparrow$)} & 
\multicolumn{1}{c}{CLIP-F ($\uparrow$)} & 
\multicolumn{1}{c}{Q-Edit ($\uparrow$)} & 
\multicolumn{1}{c|}{CM-Err ($\downarrow$)} & 
\multicolumn{1}{c}{Time ($\downarrow$)} \\ 
\midrule
VACE\cite{jiang2025vace}        & \underline{25.24} & \underline{97.80} & \underline{14.25} & \underline{3.30} & 18m 19s \\
VideoPainter\cite{bian2025videopainter} & 23.92 & 96.95 & 11.90 & 4.60 & \underline{18m 13s} \\
\midrule
ASTRA                & \textbf{27.23} & \textbf{97.93} & \textbf{14.72} & \textbf{2.83} & \textbf{10m 08s} \\
\bottomrule
\end{tabular}}
\vspace{-0.1cm}
\end{table}

\begin{table}[t]
\centering
\caption{\textbf{Cross backbone generalization.} Performance comparison of native editing pipelines versus integration with ASTRA across different mask-driven video generators. All experiments maintain identical prompts, masks, resolutions, sampling steps, and random seeds to ensure fair evaluation.}
\label{tab:backbone_agnostic}
\vspace{-0.2cm}
\renewcommand{\arraystretch}{1.08}
\setlength{\tabcolsep}{4pt} 
\resizebox{\linewidth}{!}{
\begin{tabular}{ll|c|c|c}
\toprule[0.4mm]
\rowcolor{mygray1}
\cellcolor{mygray1}\textbf{Backbone} & \cellcolor{mygray1}\textbf{Variant} & \textbf{CLIP-T $\uparrow$} & \textbf{Q-Edit $\uparrow$} & \textbf{CM-Err $\downarrow$} \\ \midrule

\multirow{2}{*}{CogVideo~\cite{hong2022cogvideo}}
& Native     & 22.19 & 11.87 & 3.46 \\
& \textbf{+ ASTRA} & \textbf{26.33} {\color[RGB]{0,150,0}\scriptsize (+18.7\%)} & \textbf{13.74} {\color[RGB]{0,150,0}\scriptsize (+15.8\%)} & \textbf{2.86} {\color[RGB]{0,150,0}\scriptsize (-17.3\%)} \\ \midrule

\multirow{2}{*}{Wan2.1~\cite{wan2025wan}}
& Native     & 24.78 & 13.24 & 3.00 \\
& \textbf{+ ASTRA} & \textbf{27.23} {\color[RGB]{0,150,0}\scriptsize (+9.9\%)} & \textbf{14.72} {\color[RGB]{0,150,0}\scriptsize (+11.2\%)} & \textbf{2.83} {\color[RGB]{0,150,0}\scriptsize (-5.7\%)} \\
\bottomrule[0.4mm]
\end{tabular}}
\vspace{-0.4cm}
\end{table}


\noindent\textbf{Cross backbone generalization.}
TABLE~\ref{tab:backbone_agnostic} evaluates the architectural generalization of ASTRA across two representative mask-driven backbones, CogVideo~\cite{hong2022cogvideo} and Wan2.1~\cite{wan2025wan}. Under strictly controlled experimental parameters (prompts, masks, resolutions, and sampling steps), integrating ASTRA yields substantial enhancements across all metrics. On CogVideo, ASTRA boosts CLIP-T from 22.19 to 26.33 and Q-Edit from 11.87 to 13.74, while reducing the layout error (CM-Err) from 3.46 to 2.86. Similarly, applying ASTRA to Wan2.1 elevates CLIP-T to 27.23 and Q-Edit to 14.72, alongside a reduction in CM-Err to 2.83. These consistent gains across distinct generative paradigms confirm that ASTRA is not constrained to a specific architecture. Instead, it effectively serves as a versatile plug-and-play conditioning layer capable of enhancing diverse mask-driven video generators.

\begin{figure*}[t]
\begin{center}
\includegraphics[width=0.95\linewidth]{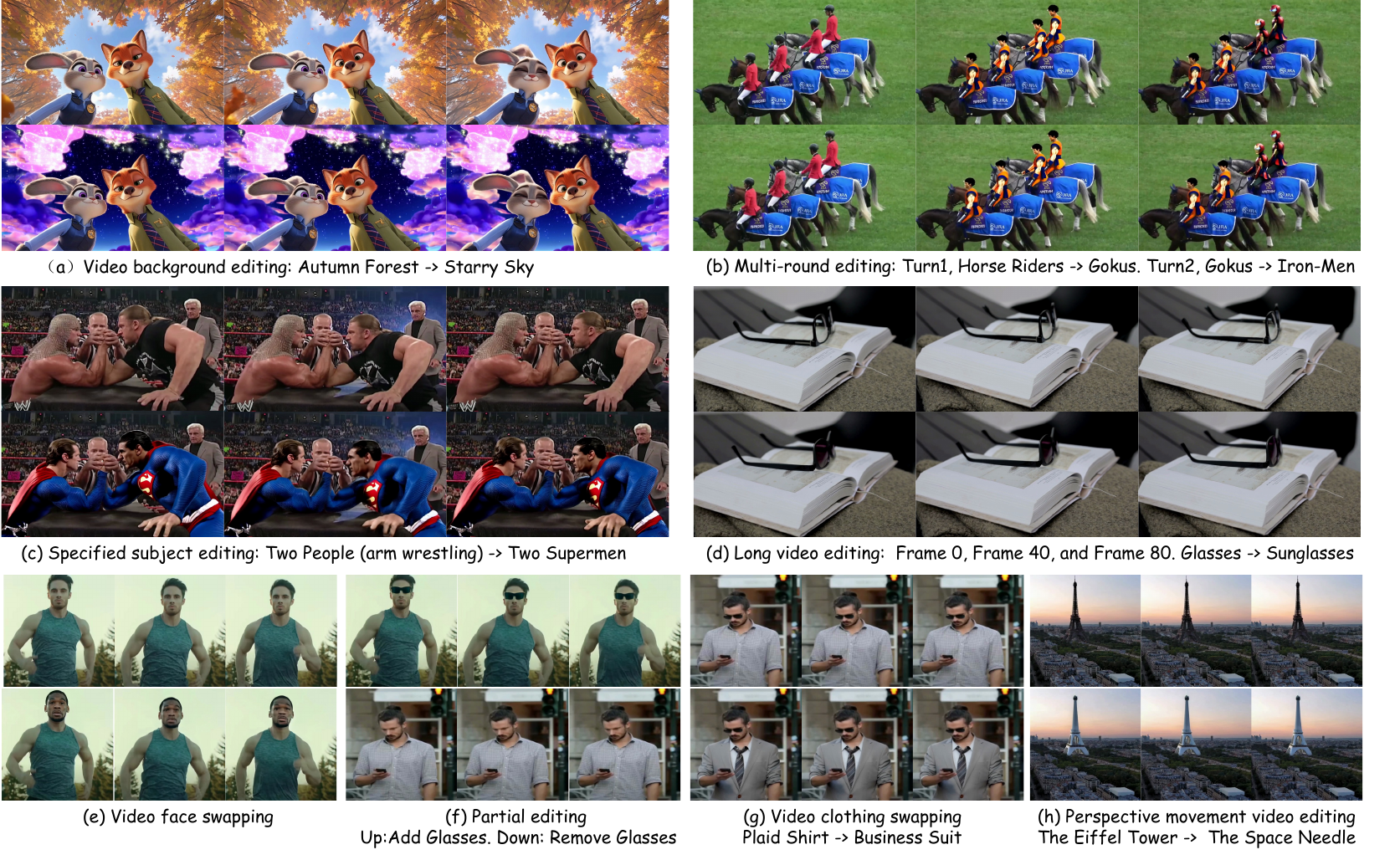}
\end{center}
\vspace{-0.3cm}
\caption{\textbf{Additional qualitative results across diverse scenarios.} Visual comparisons demonstrating the broad applicability and robust generalization of ASTRA in complex video editing tasks. More videos are available at \url{https://muzishen.github.io/ASTRA/}.}
\label{fig:more_apps} 
\vspace{-0.5cm}
\end{figure*}

\noindent\textbf{Extended Application Scenarios.}
Fig.~\ref{fig:more_apps} demonstrates the remarkable versatility of ASTRA across a wide spectrum of practical scenarios, including (a) background editing, (b) multi round editing, (c) specified subject editing, (d) long video editing, (e) face swapping, (f) partial editing, (g) clothing swapping, and (h) viewpoint change editing. These visualizations confirm that ASTRA strictly preserves non-target regions and maintains robust temporal consistency across highly complex scenes without requiring any task specific fine tuning. Furthermore, the framework seamlessly adapts to challenging structural modifications and extended temporal contexts while effectively preventing semantic drift. This extraordinary flexibility establishes ASTRA as a comprehensive and universal solution for advanced open domain video editing.

\section{Conclusion}~\label{sec:con}
We presented ASTRA, a training-free framework designed to facilitate seamless categorical transformations across an arbitrary number of subjects in video editing. To achieve this, ASTRA introduces two core components: a prompt guided multimodal alignment module and a prior based mask retargeting module. By harnessing the robust capabilities of large pretrained models, these modules generate precise multimodal conditions and temporally consistent mask sequences. This synergistic approach effectively resolves insufficient prompt side constraints and mitigates mask boundary entanglement in densely populated scenes. Crucially, ASTRA operates as a versatile plug-and-play architecture that seamlessly integrates with diverse mask-driven video generators to consistently elevate overall editing performance. Extensive evaluations on our newly established multi-subject benchmark, MSVBench, confirm that ASTRA significantly outperforms current state of the art methods.
While effective, ASTRA relies on external mask and depth priors, which may introduce localized artifacts under extreme occlusions. Additionally, the computational overhead of DiT backbones limits real-time application, motivating future work on feature distillation.




\bibliographystyle{IEEEtran}
\bibliography{refer}

\end{document}